\documentclass[10pt,twocolumn,letterpaper]{article}
\usepackage[letterpaper,margin=1in,footskip=0.45in]{geometry}
\usepackage[T1]{fontenc}
\usepackage[utf8]{inputenc}
\usepackage{times}
\usepackage{microtype}
\usepackage[table,dvipsnames]{xcolor}
\usepackage{graphicx,amsmath,amssymb,mathtools,amsthm}
\usepackage{booktabs,adjustbox,multirow,textcomp,tabularx}
\usepackage{needspace,balance}
\usepackage[numbers,sort&compress]{natbib}
\usepackage[format=plain,labelformat=simple,labelsep=period,font=small,compatibility=false]{caption}
\usepackage[font=footnotesize,skip=3pt,subrefformat=parens]{subcaption}
\definecolor{linkblue}{rgb}{0.12,0.30,0.52}
\usepackage[breaklinks,colorlinks,allcolors=linkblue]{hyperref}
\hypersetup{pdftitle={ProS2T: Progressive Fusion with Structure-Shape Transformations for Source-Free Domain Adaptation from Scratch}}
\DeclareMathSizes{8.27}{8.27}{6}{5}

\title{\LARGE ProS\textsuperscript{2}T: Progressive Fusion with Structure--Shape Transformations for Source-Free Domain Adaptation from Scratch}
\newcommand*{\PreprintAuthorNames}{Zhou Bingtao, Xiang Mian, Bingcai Chen, Qian Ning}
\author{%
  \normalsize Zhou Bingtao\textsuperscript{1,2}\quad
  Xiang Mian\textsuperscript{2}\quad
  Bingcai Chen\textsuperscript{3}\quad
  Qian Ning\textsuperscript{1}\\[5pt]
  \small\textsuperscript{1}Sichuan University\\
  \small\textsuperscript{2}Hubei Minzu University\\
  \small\textsuperscript{3}Chongqing University of Technology}
\hypersetup{pdfauthor={\PreprintAuthorNames}}

\date{}

\begin{document}
\maketitle
\begin{abstract}
Most Vision-Language (ViL)-guided works for source-free domain adaptation (SFDA) still rely on source-trained models. In those methods, we observe that different source models yield close final performance on the same target domain, suggesting the
source initialization may has limited impact. Motivated by this, we propose ViL-guided Fully-Source-Free Domain Adaptation (VF-SFDA) setting, which trains a target model from scratch using only unlabeled target data and a pretrained ViL model. Reliable ViL supervision is therefore critical, however, resizing and cropping during pre-processing can distort object shape or remove important structural cues, leading to inaccurate predictions. To address this, we propose Progressive Fusion with Structure--Shape Transformations (ProS\textsuperscript{2}T). Its S\textsuperscript{2}T module constructs complementary views: Structure Priority retains more image content, while Shape Priority largely preserves object proportions. Guided by our analysis of adaptation dynamics, ProS\textsuperscript{2}T progressively fuses each view’s ViL prediction with the same target model prediction to form two candidates, then selects the more confident one to generate hard pseudo-labels to supervise the target model, whose predictions guide ViL prompt updates. Experiments on Office-Home, VisDA-C, and DomainNet-126 demonstrate that ProS\textsuperscript{2}T achieves competitive or state-of-the-art performance compared to existing SFDA methods, despite operating under a stricter setting. Moreover, S\textsuperscript{2}T improves ViL zero-shot classification and existing ViL-guided SFDA methods as a plug-and-play module.

\end{abstract}
\section{Introduction}\label{sec:introduction}

Source-free domain adaptation (SFDA) adapts a source-trained model to an unlabeled target domain without accessing source data~\cite{li2024comprehensive,liang2020we, li2026clip}. However, it still relies on the source knowledge encoded in the model parameters.

\begin{figure}[t]
\centering
\begin{subfigure}[b]{0.48\columnwidth}
\centering\includegraphics[width=\linewidth]{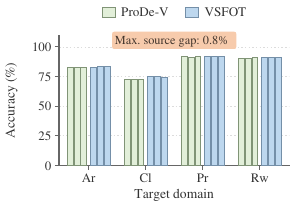}
\caption{}\label{fig:motivation_a}
\end{subfigure}\hfill
\begin{subfigure}[b]{0.50\columnwidth}
\centering\includegraphics[width=\linewidth]{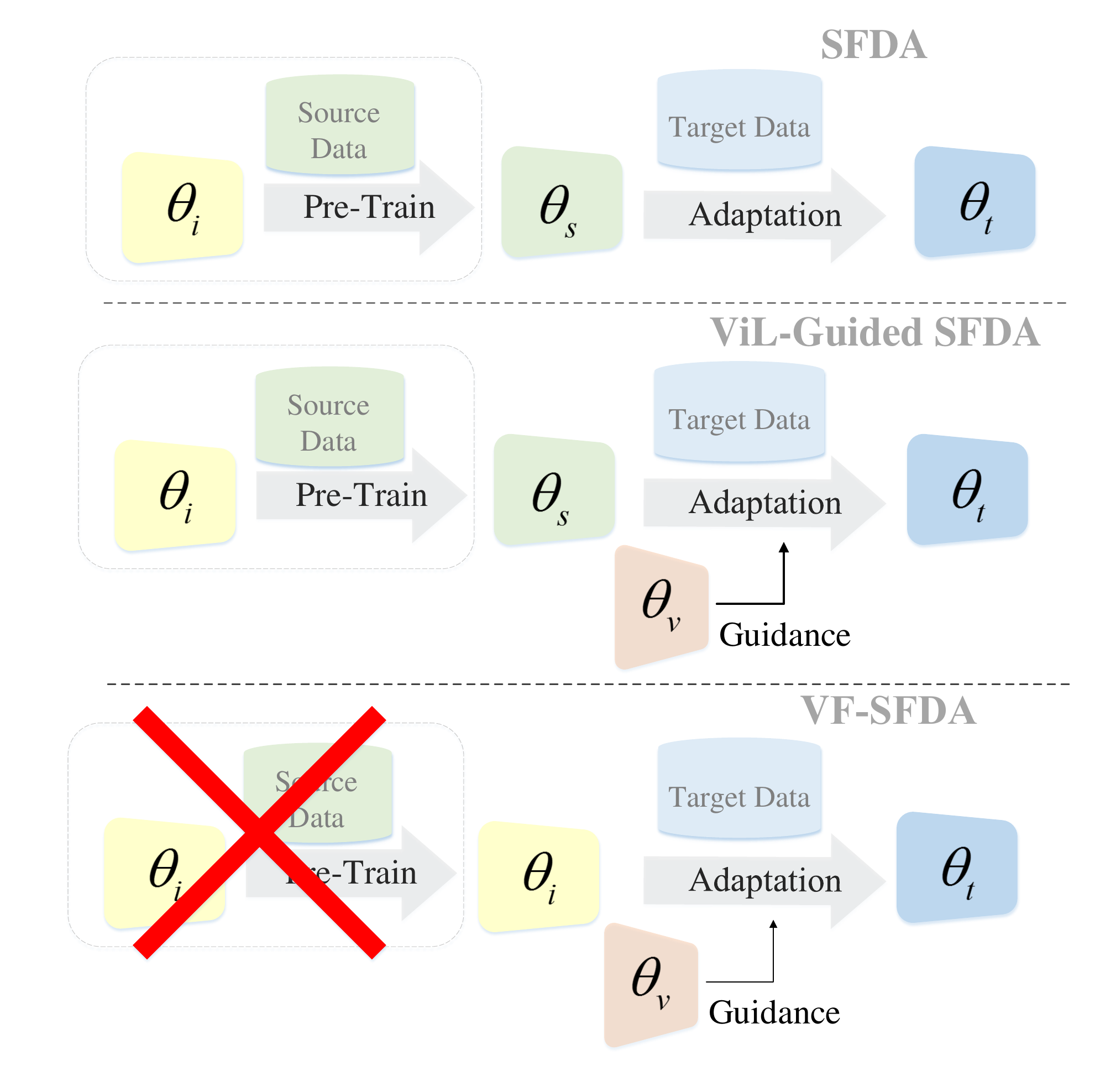}
\caption{}\label{fig:motivation_b}
\end{subfigure}
\caption{Motivation and the VF-SFDA setting. (a) Adaptation from different source models gives similar final accuracy on the same target domain under ViL guidance. (b) Comparison of conventional SFDA, ViL-guided SFDA, and VF-SFDA.}
\label{fig:motivation}
\end{figure}

Recent methods use pretrained Vision-Language (ViL) models as external guidance for adaptation and achieve substantial performance gains. For example, ProDe~\cite{ICLR2025_cd540435} corrects noisy ViL predictions through proxy denoising and uses mutual knowledge distillation for adaptation. VSFOT~\cite{Han_2026_CVPR} uses ViL semantic priors to align target features with source prototypes through optimal transport. However, we observe that, in these methods, different source initializations yield a similar final performance on the same target domain (Fig.~\ref{fig:motivation}\subref{fig:motivation_a}). This suggests that ViL guidance may reduce the influence of source initialization, raising a fundamental question: Is it possible to perform domain adaptation from scratch under ViL guidance? To explore this possibility, we introduce ViL-guided Fully-Source-Free Domain Adaptation (VF-SFDA), which trains a randomly initialized target model using a ViL model and unlabeled target data, without source data or source models, as illustrate in Fig.~\ref{fig:motivation}\subref{fig:motivation_b}. ReCLIP~\cite{Hu_2024_WACV} and DSSG~\cite{xiang2026dssg} also avoid source models but adapt ViL and retain it for inference. VF-SFDA instead uses ViL guidance to learn a standalone target model, enabling deployment without ViL.

Our analysis of VF-SFDA dynamics highlights the importance of reliable ViL supervision for learning from scratch. However, matching target images to the fixed input size of the ViL encoder~\cite{radford2021learning} requires spatial preprocessing that can alter important class cues. Resizing can distort object proportions, while cropping can remove key parts. Since recognition may rely more on the shape for some samples and on the overall structure for others, a single transformation may not preserve the most useful cues for every sample.

We therefore propose Progressive Fusion with Structure--Shape Transformations (ProS\textsuperscript{2}T), which uses a Structure--Shape Transformations (S\textsuperscript{2}T) module to improve ViL-guided supervision. S\textsuperscript{2}T pairs a Structure Priority view that retains much of the original image content with a Shape Priority view that largely preserves object proportions. The adaptation dynamics of VF-SFDA suggests that the target model’s later predictions can help improve ViL-guided supervision. We therefore progressively fuse each view’s ViL logits with the same target model logits as two candidates. The target weight starts small to limit the influence of unreliable early predictions and gradually increases to incorporate the class information learned from target data. Then we selects the more confident candidate to generate pseudo-label as distillation signal for target model. Meanwhile, the ViL model is adapted through prompt learning guided by the target model's predictions. Moreover, S\textsuperscript{2}T also consistently improves existing zero-shot ViL classification and ViL-guided SFDA methods when integrated as a plug-and-play component.

Our main contributions are summarized as follows:

\begin{itemize}

\item 
We introduce VF-SFDA, a stricter setting that removes the dependency for both source data and source models. Target adaptation relies only on a randomly initialized target model, unlabeled target data, and a ViL model.

\item 
We propose ProS\textsuperscript{2}T to improve ViL supervision through progressive fusion and complementary spatial views. Progressive fusion gradually incorporates target-model knowledge during adaptation, while S\textsuperscript{2}T provides complementary shape and structural cues to improve the quality of the pseudo-label.

\item 
Experiments on Office-Home, VisDA, and DomainNet-126 show that ProS\textsuperscript{2}T achieves performance comparable to or better than state-of-the-art SFDA methods in a more challenging setting. S\textsuperscript{2}T can serve as a plug-and-play component, consistently improving zero-shot classification across different ViL models and the performance of existing ViL-guided SFDA methods. 

\end{itemize}

\section{Related Work}\label{sec:related}

\subsection{Source-Free Domain Adaptation}\label{sec:sfda}

Conventional unsupervised domain adaptation (UDA) requires labeled source data during adaptation~\cite{long2015learning,ganin2016domain,sun2022safe,zhu2023patch,wilson2020survey}. SFDA removes this requirement, adapting a source-trained model using only unlabeled target data~\cite{liang2020we}.

SHOT \cite{liang2020we} fixes the source classifier and updates the feature extractor through information maximization and pseudo-labeling. NRC \cite{yang2021exploiting} captures local target structure by encouraging prediction consistency among reciprocal neighbors. CoWA \cite{lee2022confidence} weights target supervision by estimated confidence, while TPDS \cite{tang2024source2} searches for the target prediction distribution with explicit error control.
Recent methods further improve uncertainty control and representation learning. UCon-SFDA \cite{xu2024revisiting} controls uncertainty during adaptation, while CADTrans \cite{shao2025consistent} constructs assistant domains to learn consistent representations. ME-SFDA \cite{liu2026mesfda} uses a pyramidal memory to guide feature clustering and separate ambiguous samples. Feature Universe \cite{Lee_2026_CVPR} propagates pseudo-labels over real and virtual features and adjusts consistency regularization using feature similarity. Despite these advances, these methods retain source-trained models as the starting point for adaptation.

\subsection{Vision-Language-Guided Source-Free Domain Adaptation}\label{sec:vil_sfda}

ViL models such as CLIP \cite{radford2021learning} provide semantic priors for SFDA and can be adapted through prompt tuning \cite{zhou2022learningtoprompt,shu2022tpt}.
DIFO \cite{tang2024source} alternates between adapting ViL prompts and distilling the adapted knowledge into a source-initialized target model. DIFO++ \cite{tang2026visionlanguageprior} extends this framework through progressive distillation focused on uncertain and class-ambiguous target samples. ProDe \cite{ICLR2025_cd540435} corrects noisy ViL logits using source and target model predictions, followed by mutual knowledge distillation. DTKI \cite{zhan2026dual} transfers local structural knowledge from CLIP to support target-model learning. VSFOT \cite{Han_2026_CVPR} uses ViL semantic priors to guide optimal transport between target features and source prototypes.

ReCLIP~\cite{Hu_2024_WACV}, DSSG~\cite{xiang2026dssg}, and the SFDA extension of ImCapDA~\cite{xiang2026imcapda} also avoid task-specific source models, but retain ViL's visual backbones for inference. VF-SFDA instead trains a separate target network using ViL as an external guide during adaptation, and inference uses the trained target network alone.

\section{Method}\label{sec:method}

\subsection{Problem Formulation}\label{sec:problem}

Conventional SFDA starts from a source model $\theta_s$ pretrained on labeled source data and adapts it using unlabeled target data.

We consider the stricter VF-SFDA setting, where no source data or source-trained models are available. Adaptation uses only a randomly initialized target model $\theta_i$, unlabeled target data $\mathcal X_t=\{x_i\}_{i=1}^{N}$ with $C$ classes, and a ViL model $\theta_v$. Our goal is to adapt $\theta_i$ to the target domain using only the priors of $\theta_v$.

\subsection{ViL-Guided Progressive Fusion}
\label{sec:dynamic_process}

\begin{figure}[tbp]
\centering\includegraphics[width=\linewidth]{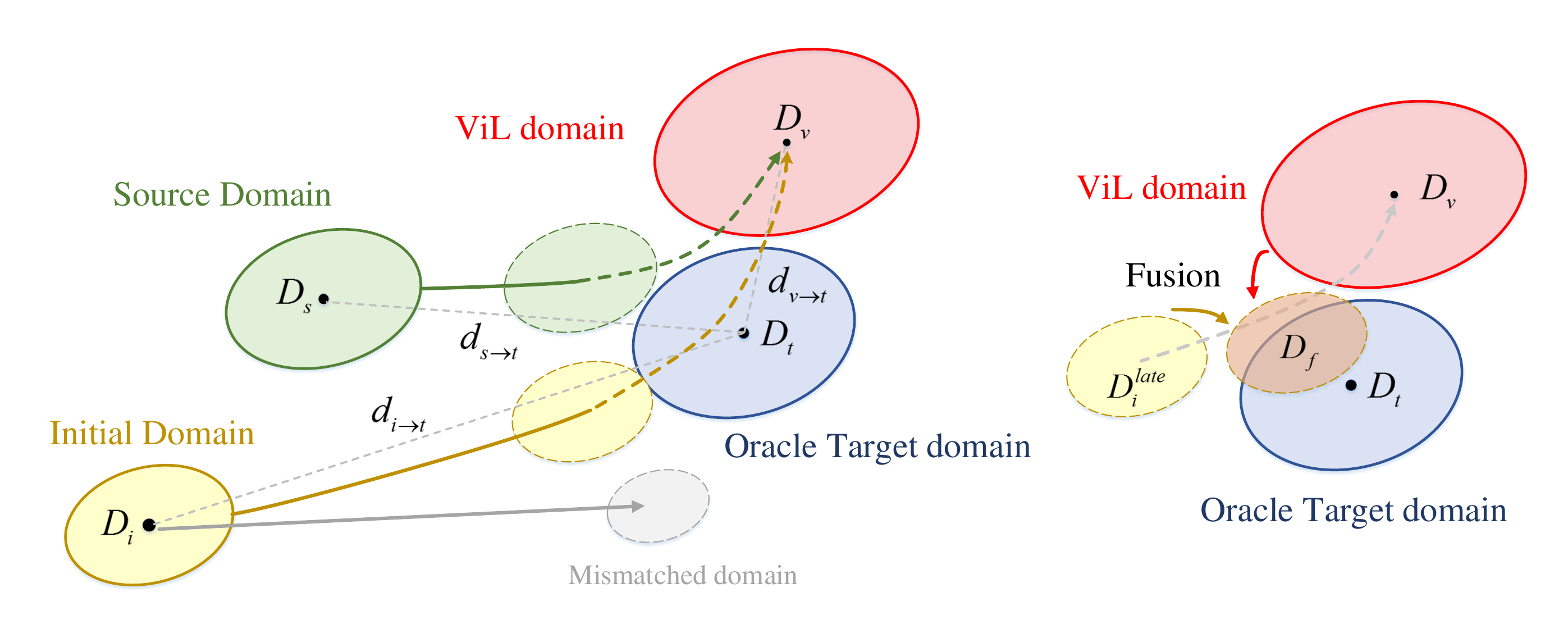}
\caption{Conceptual illustration of ViL-guided adaptation and progressive fusion. Left: ViL guides source-trained and randomly initialized models, while unreliable early supervision may lead to a mismatched predictive state. Right: late-stage target predictions complement ViL predictions, motivating their fusion to improve target-domain supervision.}
\label{fig:dynamic_process}
\end{figure}

The left part of Fig.~\ref{fig:dynamic_process} illustrates the dynamics of VF-SFDA. \(D_s\) and \(D_i\) denote the initial predictive states of source-trained and randomly initialized models, while \(D_v\) and \(D_t\) represent ViL and oracle predictions on the same target samples. 
ViL models often provide more accurate initial predictions than unadapted source models. For example,ProDe reports 76.1\% zero-shot accuracy for CLIP on Office-Home, compared with 59.2\% for unadapted source models \cite{ICLR2025_cd540435}.
We therefore rely mainly on ViL supervision early in training to guide the randomly initialized model toward the target domain. This helps reduce the risk of drifting toward a mismatched predictive state.

The right part illustrates the motivation for fusion later in training. 
ViL supervision and target data objectives, such as batch-level class balancing, jointly shape the \(D_i\)'s learning. 
As adaptation progresses, its late-stage predictions \(D_i^{\mathrm{late}}\) may provide target-specific information that complements \(D_v\). 
This motivates combining the two sources to construct fused supervision \(D_f\), with the aim of bringing predictions closer to the oracle state \(D_t\).

Based on this analysis, we adopt progressive logit fusion to prioritize ViL supervision early in training and gradually increase the contribution of \(D_i\)'s predictions as their reliability improves.
For the sample \(x_i\), the target and ViL models produce logits:
\begin{equation}
\mathbf z_i
=
\theta_i\!\left(x_i\right),
\qquad
\mathbf z_{v,i}
=
\theta_v\!\left(x_i\right).
\label{eq:target_logits}
\end{equation}

To account for differences in output scales, we standardize each logit vector across classes:
\begin{equation}
\mathcal N(\mathbf z)
=
\frac{\mathbf z-\mu(\mathbf z)}
{\sigma(\mathbf z)+\epsilon},
\label{eq:logit_standardization}
\end{equation}
where \(\mu(\mathbf z)\) and \(\sigma(\mathbf z)\) are the mean and standard deviation across classes, and \(\epsilon>0\) ensures numerical stability. Following ONE's multi-branch fusion strategy~\cite{lan2018knowledge}, we treat $\theta_i$ and $\theta_v$ as two prediction branches and fuse their standardized logits:

\begin{equation}
\mathbf z_{f,i}
=
\mathcal N\!\left(\mathbf z_{v,i}\right)
+
\lambda(t)\mathcal N(\mathbf z_i).
\label{eq:progressive_fusion}
\end{equation}

Under ViL guidance, \(D_i\)'s predictions can quickly approach ViL predictions early in the training, but remain unreliable. 
We therefore use a cubic schedule to keep their early contribution small and increase it more rapidly later:
\begin{equation}
\lambda(t)
=
\min\left\{
1,
\left(\frac{t}{T_r}\right)^3
\right\},
\qquad t\geq 0,
\label{eq:fusion_schedule}
\end{equation}
where \(t\) denotes the training progress measured in epochs, starting from zero, and \(T_r>0\) controls the duration of the weight increase.

As this process illustrates, ViL predictions directly guide the learning of $\theta_i$, making their quality critical to adaptation and motivating the Structure--Shape Transformations introduced next.

\begin{figure}[t]
\centering
\begin{subfigure}{\columnwidth}
\centering\includegraphics[width=0.85\linewidth]{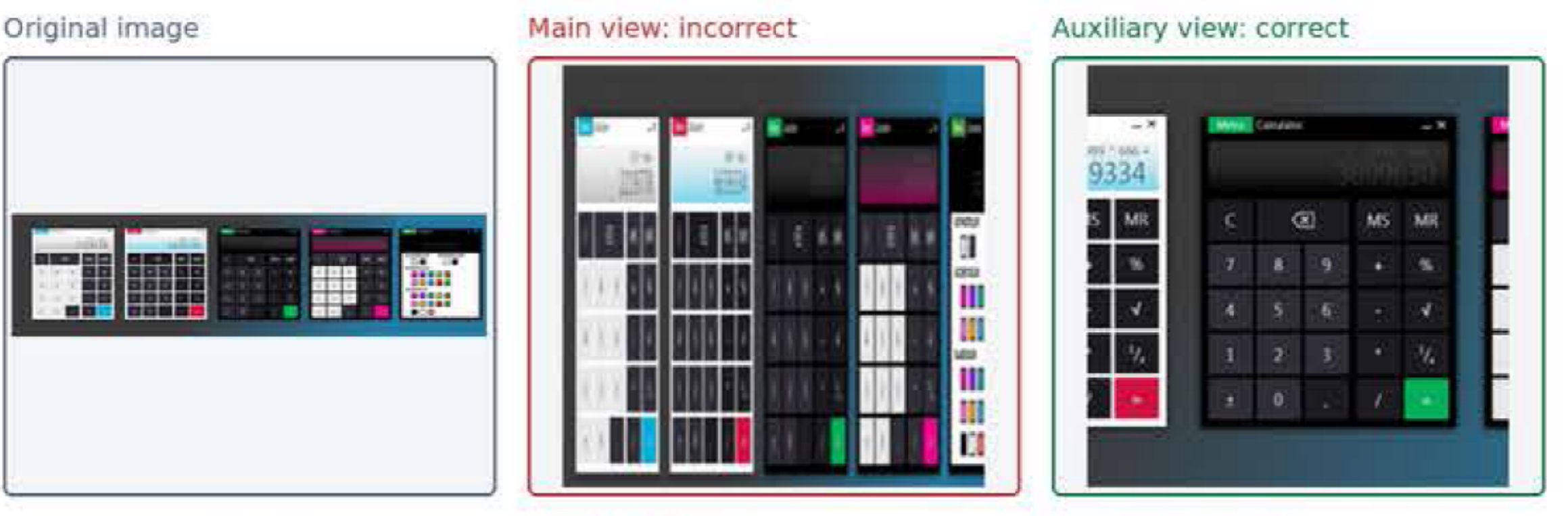}
\caption{}\label{fig:case_calculator}
\end{subfigure}
\par\smallskip
\begin{subfigure}{\columnwidth}
\centering\includegraphics[width=0.85\linewidth,trim=0 325 0 100,clip]{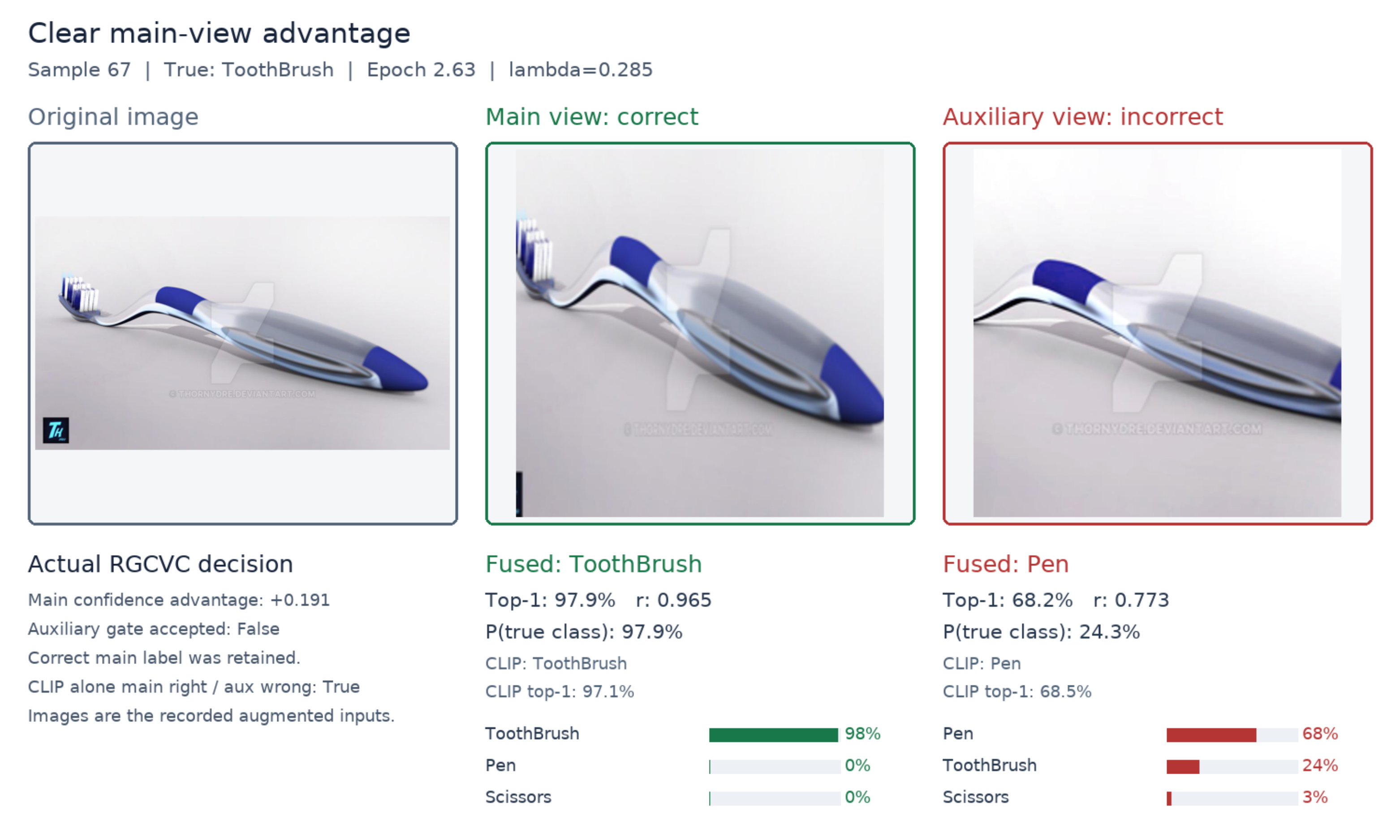}
\caption{}\label{fig:case_toothbrush}
\end{subfigure}
\caption{Effects of spatial transformations on ViL predictions. Each panel shows the original image, the main view, and the auxiliary view. (a) The main view misclassifies a calculator as a calendar. (b) The auxiliary view misclassifies a toothbrush as a pen.}
\label{fig:spatial_cases}
\end{figure}

\subsection{Structure--Shape Transformations (S\textsuperscript{2}T)}\label{sec:csvc}

\begin{figure*}[t]
\centering\includegraphics[width=\textwidth]{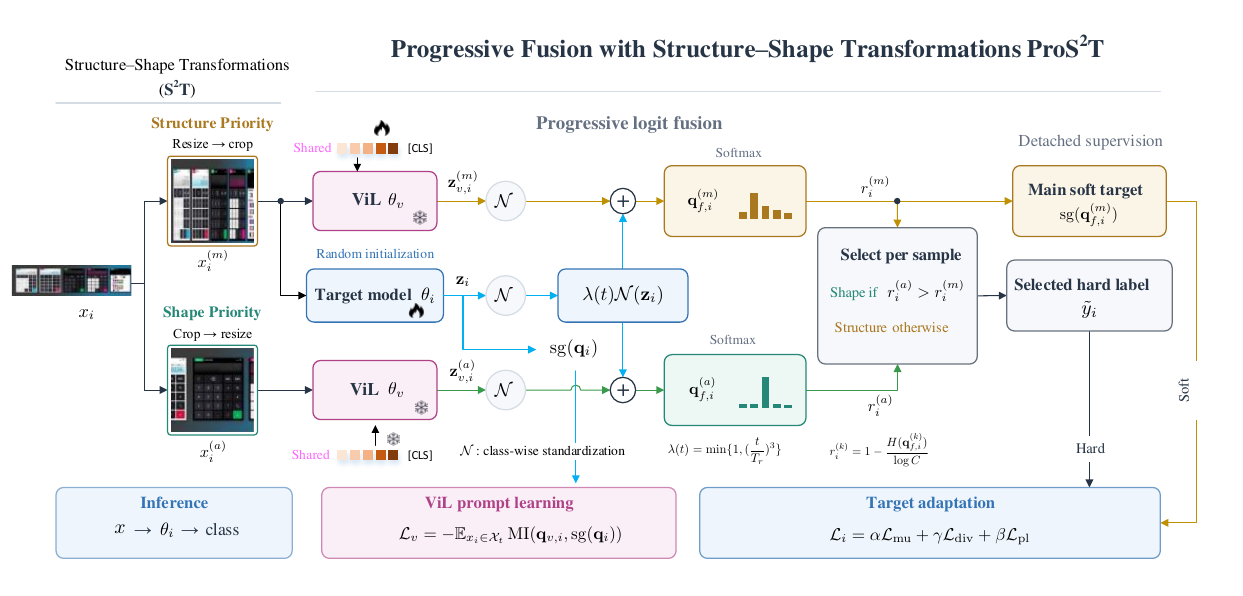}
\caption{Overview of ProS\textsuperscript{2}T. S\textsuperscript{2}T uses complementary Structure Priority and Shape Priority views to improve ViL supervision. Progressive fusion gradually incorporates target-model predictions, and confidence-based selection refines pseudo-labels. The randomly initialized target model learns from this supervision and guides ViL prompt updates.}
\label{fig:framework}
\end{figure*}

Images vary in resolution and aspect ratio.
Spatial transformations are therefore required to match the ViL model's input size.
For example, a \(1080\times720\) image is converted to \(224\times224\) before entering the CLIP visual encoder.
This process can change object shape, scale, and visible content, affecting the cues available for recognition.

We define that the main transformation \(\mathcal T_m\) follows the resize-then-crop pipeline~\cite{liang2020we} commonly used in SFDA:
\begin{equation}
\mathcal T_m(x_i)
=
\mathcal P\!\left(
\mathcal F\!\left(
\mathcal C_{\mathrm{rand}}\!\left(
\mathcal R_{\mathrm{sq}}(x_i)
\right)\right)\right),
\label{eq:main_transform}
\end{equation}
where \(\mathcal R_{\mathrm{sq}}\) resizes the entire image to a square,
and \(\mathcal C_{\mathrm{rand}}\) takes a random crop matching the ViL input size.
\(\mathcal F\) denotes random horizontal flipping, and \(\mathcal P\) converts the image to a normalized tensor.
We describe this as \textbf{Structure Priority}: resizing retains much of the original content.
However, it can distort object proportions.
As shown in Fig.~\ref{fig:spatial_cases}\subref{fig:case_calculator}, \(\mathcal T_m\) stretches the calculator's display and keys vertically, distorting their proportions and leading to an incorrect prediction: calendar.

To reduce such distortion, we introduce an auxiliary transformation \(\mathcal T_a\) with \textbf{Shape Priority}:
\begin{equation}
\mathcal T_a(x_i)
=
\mathcal P\!\left(
\mathcal F\!\left(
\mathcal R_{\mathrm{in}}\!\left(
\mathcal C_{\mathrm{ratio}}(x_i)
\right)\right)\right),
\label{eq:aux_transform}
\end{equation}
where \(\mathcal C_{\mathrm{ratio}}\) first samples a crop with an aspect ratio close to that of the model input.
\(\mathcal R_{\mathrm{in}}\) then resizes it to the required input size, largely preserving object proportions and limiting geometric distortion.
In Fig.~\ref{fig:spatial_cases}\subref{fig:case_calculator}, the auxiliary view presents the calculator with more natural proportions and produces a correct prediction.

However, cropping can also remove key class cues.
Fig.~\ref{fig:spatial_cases}\subref{fig:case_toothbrush} shows the opposite case: the main view retains the toothbrush bristles and correctly identifies the toothbrush.
The auxiliary view removes the bristle region, and misclassifies it as a pen.

These examples show that shape preservation and visible content jointly affect recognition, and either view may lose useful cues for a given sample.
We therefore retain both transformations to construct complementary observations:
\begin{equation}
x_i^{(m)}
=
\mathcal T_m(x_i),
\qquad
x_i^{(a)}
=
\mathcal T_a(x_i).
\label{eq:views}
\end{equation}

The ViL model produces the corresponding prediction logits:
\begin{equation}
\begin{aligned}
\mathbf z_{v,i}^{(m)}
=
\theta_v\!\left(x_i^{(m)}\right),
\qquad
\mathbf z_{v,i}^{(a)}
=
\theta_v\!\left(x_i^{(a)}\right).
\end{aligned}
\label{eq:vil_views}
\end{equation}

These view-specific predictions provide candidates for the supervision signals described next.

\subsection{Progressive Fusion with Structure-Shape Transformations (ProS\textsuperscript{2}T)}
\label{sec:refinement}

Fig.~\ref{fig:framework} summarizes ProS\textsuperscript{2}T. We apply the progressive fusion defined in Sec.~\ref{sec:dynamic_process} to the two S\textsuperscript{2}T views, combining its standardized ViL logits with the same target-model logits from the main view:
\begin{equation}
\begin{aligned}
\mathbf z_{f,i}^{(m)}
=
\mathcal N\!\left(\mathbf z_{v,i}^{(m)}\right)
+
\lambda(t)\mathcal N(\mathbf z_i),
\\
\mathbf z_{f,i}^{(a)}
=
\mathcal N\!\left(\mathbf z_{v,i}^{(a)}\right)
+
\lambda(t)\mathcal N(\mathbf z_i).
\end{aligned}
\label{eq:two_fusions}
\end{equation}

Their class distributions are:
\begin{equation}
\mathbf q_{f,i}^{(k)}
=
\operatorname{softmax}
\!\left(\mathbf z_{f,i}^{(k)}\right),
\qquad k\in\{m,a\}.
\label{eq:candidate_probs}
\end{equation}

We select the candidate with the highest overall confidence, computed from the fused predictions of the ViL and target models. 
Lower predictive entropy indicates higher confidence~\cite{grandvalet2004semi}. We therefore use its normalized form to compare the two fused candidates for each sample:
\begin{equation}
r_i^{(k)}
=
1-
\frac{H\!\left(\mathbf q_{f,i}^{(k)}\right)}{\log C},
\qquad k\in\{m,a\},
\label{eq:reliability}
\end{equation}
where \(C\) is the number of classes and
\begin{equation}
H\!\left(\mathbf q_{f,i}^{(k)}\right)
=
-\sum_{c=1}^{C}
q_{f,i,c}^{(k)}
\log q_{f,i,c}^{(k)}.
\label{eq:entropy}
\end{equation}

A higher reliability score indicates a more concentrated class distribution. 
The candidate hard labels are:
\begin{equation}
\hat y_i^{(k)}
=
\underset{c\in\{1,\ldots,C\}}{\arg\max}
\;
q_{f,i,c}^{(k)},
\qquad k\in\{m,a\}.
\label{eq:hard_candidates}
\end{equation}

We retain the main view label by default and replace it only when the auxiliary candidate has a higher reliability score:
\begin{equation}
\tilde y_i
=
\begin{cases}
\hat y_i^{(a)},
&
r_i^{(a)}>r_i^{(m)},
\\[2mm]
\hat y_i^{(m)},
&
\text{otherwise}.
\end{cases}
\label{eq:correction}
\end{equation}

The selected label \(\tilde y_i\) provides hard supervision, while the main candidate \(\mathbf q_{f,i}^{(m)}\) provides soft supervision. 
Both candidate distributions are detached when used as supervision.

\subsection{Target Adaptation and ViL Prompt Updating}\label{sec:optimization}

\subsubsection{Target Model Adaptation}\label{sec:target}

We train the target model with three objectives: mutual-information alignment, class-distribution balancing, and hard pseudo-label supervision. Let the target-model prediction and the main-view fused prediction be

\begin{equation}
\begin{aligned}\mathbf q_i&=\operatorname{softmax}(\mathbf z_i),
\qquad
\mathbf q_{f,i}^{(m)}
=
\operatorname{softmax}\!\left(
\operatorname{sg}(\mathbf z_{f,i}^{(m)})
\right)\end{aligned}
\label{eq:target_probs}
\end{equation}

where \(\operatorname{sg}(\cdot)\) denotes stop-gradient. Following IIC~\cite{ji2019iic} and its use in ViL-guided SFDA methods such as DIFO and ProDe~\cite{tang2024source,ICLR2025_cd540435}, we maximize the mutual information between the fused prediction and the target-model prediction:

\begin{equation}
\mathcal L_{\mathrm{mu}}
=
-\mathbb E_{x_i\in\mathcal X_t}
\operatorname{MI}(\mathbf q_{f,i}^{(m)},\mathbf q_i).
\label{eq:mi}
\end{equation}

Here, \(\operatorname{MI}\) is estimated from the paired predictions in each minibatch. This objective transfers soft supervision from progressive fusion to the target model.

To discourage prediction collapse, we maximize the entropy of the batch-averaged class distribution~\cite{gomes2010discriminative}. We define

\begin{equation}
\mathcal L_{\mathrm{div}}
=
\sum_{c=1}^{C}\bar q_c\log\bar q_c.
\label{eq:div}
\end{equation}

Where \(\bar q_c=\frac{1}{B}\sum_{i=1}^{B}q_{i,c}\), and \(B\) is the size of a minibatch.

We also supervise the target model with the corrected hard pseudo-label \(\tilde y_i\) from confidence-based selection:

\begin{equation}
\mathcal L_{\mathrm{pl}}
=
-\mathbb E_{x_i\sim\mathcal X_t}
\left[
\sum_{c=1}^{C}
\tilde y_{i,c}\log q_{i,c}
\right],
\label{eq:ce}
\end{equation}

\(\tilde y_{i,c}\) is the \(c\)-th component of the one-hot encoding of the refined pseudo-label \(\tilde y_i\). The overall target-model objective is:

\begin{equation}
\mathcal L_i
=
\alpha\mathcal L_{\mathrm{mu}}
+
\gamma\mathcal L_{\mathrm{div}}
+
\beta\mathcal L_{\mathrm{pl}},
\label{eq:total}
\end{equation}

where \(\alpha\), \(\gamma\), and \(\beta\) weight the three terms.

\subsubsection{ViL Prompt Updating}\label{sec:prompt}

Following the paradigm in recent ViL-guided SFDA methods~\cite{tang2024source,ICLR2025_cd540435,tang2026visionlanguageprior}, we update the ViL model's learnable context vectors while keeping the remaining ViL parameters frozen. The main view ViL prediction is

\begin{equation}
\mathbf q_{v,i}
=
\operatorname{softmax}\!\left(\mathbf z_{v,i}^{(m)}\right).
\label{eq:vil_probs}
\end{equation}

We update the prompts by maximizing the mutual information between the ViL and target-model predictions:

\begin{equation}
\mathcal L_v
=
-\mathbb E_{x_i\in\mathcal X_t}
\operatorname{MI}\!\left(
\mathbf q_{v,i},
\operatorname{sg}(\mathbf q_i)
\right).
\label{eq:prompt}
\end{equation}

During this update, the target-model prediction is detached, so gradients update only the learnable context vectors.

\section{Experiments}\label{sec:experiments}

\subsection{Experimental Setup}\label{sec:setup}
\begin{table*}[t]
\centering
\renewcommand{\arraystretch}{0.8}
\scriptsize
\caption{Closed-set accuracy (\%) on \textbf{Office-Home} and \textbf{VisDA} (mean per-class). SMF: source-model-free. \textbf{bold}: best.}
\label{tab:office_home_visda}
\label{tab:office_home_close-set}
\begin{adjustbox}{width=\textwidth}
\begin{tabular}{lcccccccc}
\toprule
\multirow{3}{*}{Method} & \multirow{3}{*}{SMF} & \multirow{3}{*}{Venue} & \multicolumn{5}{c}{Office-Home} & \multirow{3}{*}{\shortstack{VisDA\\Mean}} \\
\cmidrule(lr){4-8}
& & & Ar & Cl & Pr & Rw & \multirow{2}{*}{Avg.} & \\
\cmidrule(lr){4-4} \cmidrule(lr){5-5} \cmidrule(lr){6-6} \cmidrule(lr){7-7}
& & & Cl\, Pr\, Rw$\to$Ar & Ar\, Pr\, Rw$\to$Cl & Ar\, Cl\, Rw$\to$Pr & Ar\, Cl\, Pr$\to$Rw & & \\
\midrule
SHOT \cite{liang2020we} & \texttimes & ICML20 & 68.0\, 67.9\, 74.2 & 56.7\, 54.5\, 58.6 & 77.9\, 78.0\, 84.5 & 80.6\, 79.4\, 82.3 & 71.9 & 82.7 \\
NRC \cite{yang2021exploiting} & \texttimes & NeurIPS21 & 68.1\, 65.3\, 71.0 & 57.7\, 56.4\, 58.6 & 80.3\, 79.8\, 85.6 & 82.0\, 78.6\, 83.0 & 72.2 & 85.9 \\
AaD \cite{yang2022attracting} & \texttimes & NeurIPS22 & 68.9\, 67.2\, 72.1 & 59.3\, 57.4\, 58.5 & 79.3\, 79.8\, 85.4 & 82.1\, 79.5\, 83.1 & 72.7 & 88.0 \\
AdaCon \cite{chen2022contrastive} & \texttimes & CVPR22 & -- & -- & -- & -- & -- & 86.8 \\
CoWA \cite{lee2022confidence} & \texttimes & ICML22 & 69.1\, 67.7\, 72.8 & 56.9\, 57.2\, 60.5 & 78.4\, 80.0\, 84.5 & 81.0\, 79.9\, 82.4 & 72.5 & 86.9 \\
PLUE \cite{litrico2023guiding} & \texttimes & CVPR23 & -- & -- & -- & -- & -- & 88.3 \\
TPDS \cite{tang2024source2} & \texttimes & IJCV24 & 70.6\, 69.8\, 74.5 & 59.3\, 56.8\, 61.2 & 80.3\, 79.4\, 85.3 & 82.1\, 80.9\, 82.1 & 73.5 & 87.6 \\
CADTrans \cite{shao2025consistent} & \texttimes & TIP25 & 83.1\, 80.1\, 81.8 & 70.3\, 62.9\, 74.3 & 88.7\, 90.2\, 92.5 & 90.0\, 89.9\, 90.5 & 82.9 & -- \\
UCon-SFDA \cite{xu2024revisiting} & \texttimes & ICLR25 & 69.3\, 67.0\, 72.2 & 61.5\, 62.2\, 61.9 & 80.5\, 80.2\, 85.5 & 82.1\, 78.7\, 82.0 & 73.6 & 89.6 \\
ME-SFDA \cite{liu2026mesfda} & \texttimes & AAAI26 & 69.2\, 66.7\, 74.8 & 61.3\, 58.9\, 60.9 & 80.8\, 81.3\, 86.7 & 82.9\, 80.8\, 83.1 & 74.0 & 89.6 \\
Feature Universe \cite{Lee_2026_CVPR} & \texttimes & CVPR26 & 71.2\, 70.3\, 74.7 & 59.7\, 58.7\, 63.1 & 81.4\, 80.4\, 85.7 & 82.8\, 81.3\, 83.5 & 74.4 & 87.7 \\
\midrule
CLIP (zero-shot) \cite{radford2021learning} & \textbf{\checkmark} & ICML21 & 75.2 & 60.2 & 84.2 & 85.5 & 76.3 & 85.3 \\
DIFO-V \cite{tang2024source} & \texttimes & CVPR24 & 82.5\, 80.9\, 83.4 & 70.6\, 70.1\, 70.5 & 90.6\, 90.6\, 91.2 & 88.8\, 88.8\, 88.9 & 83.1 & 90.3 \\
ProDe-V \cite{ICLR2025_cd540435} & \texttimes & ICLR25 & 82.5\, 82.5\, 83.0 & 72.7\, 72.5\, 72.6 & 92.3\, 91.5\, 92.2 & 90.5\, 90.7\, 90.8 & 84.5 & 91.0 \\
DUET \cite{lee2025duet} & \texttimes & NeurIPS25 & 83.6\, 82.7\, 83.6 & 73.6\, 73.7\, 74.0 & 90.4\, 90.7\, 91.2 & 91.0\, 90.9\, 91.2 & 84.7 & \textbf{91.4} \\
DTKI \cite{zhan2026dual} & \texttimes & IPM26 & 82.3\, 82.8\, 82.7 & 72.0\, 71.1\, 71.1 & 91.7\, 91.9\, 91.6 & 90.0\, 90.0\, 90.3 & 84.0 & 90.8 \\
VSFOT \cite{Han_2026_CVPR} & \texttimes & CVPR26 & 83.0\, 83.5\, 83.7 & 75.0\, 75.0\, 74.3 & 91.7\, 91.8\, 91.9 & \textbf{91.4}\, \textbf{91.4}\, \textbf{91.4} & 85.3 & 90.6 \\
DIFO++ \cite{tang2026visionlanguageprior} & \texttimes & IJCV26 & 83.6\, 80.8\, 84.3 & 73.2\, 73.8\, 74.2 & 91.4\, 91.0\, 92.3 & 89.7\, 89.8\, 90.0 & 84.5 & 90.5 \\
\midrule
\rowcolor{gray!20} CLIP (zero-shot) + S\textsuperscript{2}T & \textbf{\checkmark} & -- & 77.8 & 64.1 & 86.5 & 87.5 & 79.0\,(+2.7) & 85.8\,(+0.5) \\
\rowcolor{gray!20} DIFO-V + S\textsuperscript{2}T & \texttimes & -- & 83.7\, 82.8\, 84.1 & 73.4\, 73.0\, 72.9 & 91.1\, 91.0\, 91.7 & 90.3\, 90.1\, 90.0 & 84.5\,(+1.4) & 90.9\,(+0.6) \\
\rowcolor{gray!20} ProDe-V + S\textsuperscript{2}T & \texttimes & -- & \textbf{84.4}\, 84.2\, 84.1 & 75.0\, 75.1\, \textbf{75.2} & 92.4\, 92.4\, 92.4 & 90.9\, 91.1\, 91.3 & \textbf{85.7}\,(+1.2) & 91.0\,(+0.0) \\
\midrule
\rowcolor{gray!20} ProS\textsuperscript{2}T (Ours) & \textbf{\checkmark} & -- & 83.5 & 74.4 & \textbf{92.8} & 90.8 & 85.4 & 91.1 \\
\bottomrule
\end{tabular}
\end{adjustbox}
\end{table*}

\subsubsection{Datasets}\label{sec:datasets}

We evaluate our method on three benchmarks: 

Office-Home~\cite{venkateswara2017officehome} contains about 15,500 images from 65 categories across four domains: Artistic (Ar), Clipart (Cl), Product (Pr), and Real-World (Rw). The domains cover distinct visual styles.

VisDA-C~\cite{peng2017visda} is a synthetic-to-real domain adaptation benchmark with 12 object categories. It contains about 152,000 synthetic images and 55,000 real images. We use the real images as the target domain.

DomainNet-126~\cite{peng2019domainnet,saito2019mme} is a subset of DomainNet with about 145,000 images from 126 categories across four domains: Clipart (C), Painting (P), Real (R), and Sketch (S).

\subsubsection{Comparison Methods and Evaluation Protocol}\label{sec:baselines}

We report closed-set classification accuracy on Office-Home and DomainNet-126, and mean per-class accuracy on VisDA-C. We compare our method with representative SFDA and ViL-guided SFDA methods.
The SFDA baselines include SHOT~\cite{liang2020we},
NRC~\cite{yang2021exploiting},
AaD~\cite{yang2022attracting},
AdaCon~\cite{chen2022contrastive},
CoWA~\cite{lee2022confidence},
PLUE~\cite{litrico2023guiding},
TPDS~\cite{tang2024source2},
UCon-SFDA~\cite{xu2024revisiting},
CADTrans~\cite{shao2025consistent},
ME-SFDA~\cite{liu2026mesfda},
and Feature Universe~\cite{Lee_2026_CVPR}.
The ViL-guided baselines include DIFO-V~\cite{tang2024source},
ProDe-V~\cite{ICLR2025_cd540435},
DUET~\cite{lee2025duet},
DTKI~\cite{zhan2026dual},
VSFOT~\cite{Han_2026_CVPR},
and DIFO++~\cite{tang2026visionlanguageprior}.
We also report CLIP zero-shot accuracy~\cite{radford2021learning} as a reference.

We further evaluate S\textsuperscript{2}T as a plug-and-play module in CLIP zero-shot, DIFO-V, and ProDe-V. We apply S\textsuperscript{2}T to the ViL branch. For CLIP zero-shot, selection is applied directly to the two ViL predictions; for DIFO-V and ProDe-V, S\textsuperscript{2}T refines only hard pseudo-labels. Our reproductions of DIFO-V and ProDe-V achieve accuracy comparable to the results reported in the original papers.

\subsection{Comparison with State-of-the-Art Methods}\label{sec:comparison}

\begin{table*}[t]
\centering
\renewcommand{\arraystretch}{0.8}
\scriptsize
\caption{Closed-set accuracy (\%) on \textbf{DomainNet-126}. SMF: source-model-free. 
 \textbf{bold}: best.}
\label{tab:domainnet126}
\begin{adjustbox}{width=\textwidth}
\begin{tabular}{lccccccc}
\toprule
\multirow{2}{*}{Method} & \multirow{2}{*}{SMF} & \multirow{2}{*}{Venue} & C & P & R & S & \multirow{2}{*}{Avg.} \\
\cmidrule(lr){4-4} \cmidrule(lr){5-5} \cmidrule(lr){6-6} \cmidrule(lr){7-7}
& & & P\, R\, S$\to$C & C\, R\, S$\to$P & C\, P\, S$\to$R & C\, P\, R$\to$S & \\
\midrule
SHOT \cite{liang2020we} & \texttimes & ICML20 & 67.9\, 67.7\, 70.2 & 63.5\, 67.6\, 64.0 & 78.2\, 81.3\, 78.0 & 59.5\, 61.7\, 57.8 & 68.1 \\
NRC \cite{yang2021exploiting} & \texttimes & NeurIPS21 & 62.9\, 64.7\, 69.4 & 62.6\, 69.4\, 65.8 & 77.1\, 81.3\, 78.7 & 58.3\, 60.7\, 58.7 & 67.5 \\
AdaCon \cite{chen2022contrastive} & \texttimes & CVPR22 & 62.2\, 63.1\, 67.1 & 60.8\, 68.1\, 66.0 & 74.8\, 78.3\, 75.4 & 55.9\, 58.2\, 55.6 & 65.4 \\
CoWA \cite{lee2022confidence} & \texttimes & ICML22 & 66.2\, 69.0\, 69.0 & 64.6\, 67.2\, 65.8 & 80.6\, 79.8\, 79.9 & 60.6\, 60.8\, 60.0 & 68.6 \\
PLUE \cite{litrico2023guiding} & \texttimes & CVPR23 & 61.6\, 61.6\, 67.5 & 59.8\, 65.9\, 64.3 & 74.0\, 78.5\, 76.0 & 56.0\, 57.9\, 53.8 & 64.7 \\
TPDS \cite{tang2024source2} & \texttimes & IJCV24 & 65.6\, 66.4\, 68.6 & 62.9\, 67.0\, 64.3 & 77.1\, 79.0\, 75.3 & 59.8\, 61.5\, 58.2 & 67.1 \\
CADTrans \cite{shao2025consistent} & \texttimes & TIP25 & 68.0\, 64.1\, 73.5 & 75.4\, 74.5\, 76.3 & 89.2\, 89.3\, 88.9 & 66.5\, 63.1\, 58.4 & 73.9 \\
Feature Universe \cite{Lee_2026_CVPR} & \texttimes & CVPR26 & 75.4\, 76.4\, 76.5 & 66.2\, 72.4\, 70.3 & 80.6\, 83.7\, 80.6 & 64.8\, 69.5\, 67.0 & 73.6 \\
\midrule
CLIP (zero-shot) \cite{radford2021learning} & \textbf{\checkmark} & ICML21 & 77.3 & 76.2 & 88.6 & 71.1 & 78.3 \\
DIFO-V \cite{tang2024source} & \texttimes & CVPR24 & 80.0\, 80.8\, 80.5 & 76.6\, 77.3\, 76.7 & 87.2\, 87.4\, 87.3 & 74.9\, 75.6\, 75.5 & 80.0 \\
ProDe-V \cite{ICLR2025_cd540435} & \texttimes & ICLR25 & 85.0\, 85.5\, 85.5 & 83.2\, 83.1\, 83.4 & 92.4\, 92.3\, 92.4 & 79.0\, 79.3\, 79.1 & 85.0 \\
DUET \cite{lee2025duet} & \texttimes & NeurIPS25 & 82.4\, 82.8\, 83.0 & 80.1\, 80.6\, 80.3 & 89.6\, 89.8\, 89.6 & 79.0\, 79.2\, 78.8 & 82.9 \\
DTKI \cite{zhan2026dual} & \texttimes & IPM26 & 82.4\, 80.7\, 82.3 & 77.5\, 78.7\, 78.9 & 88.9\, 88.0\, 88.3 & 76.1\, 76.9\, 74.6 & 81.1 \\
VSFOT \cite{Han_2026_CVPR} & \texttimes & CVPR26 & 83.0\, 83.3\, 83.5 & 79.5\, 79.8\, 79.6 & 90.4\, 90.5\, 90.4 & 77.2\, 77.5\, 76.9 & 82.6 \\
DIFO++ \cite{tang2026visionlanguageprior} & \texttimes & IJCV26 & 81.7\, 82.3\, 82.3 & 78.1\, 80.2\, 79.4 & 89.7\, 89.6\, 89.7 & 76.5\, 77.2\, 75.9 & 81.9 \\
\midrule
\rowcolor{gray!20} CLIP (zero-shot) + S\textsuperscript{2}T & \textbf{\checkmark} & -- & 79.9 & 78.6 & 90.0 & 74.6 & 80.8\,(+2.5) \\
\rowcolor{gray!20} DIFO-V + S\textsuperscript{2}T & \texttimes & -- & 84.4\, 84.9\, 84.6 & 81.9\, 82.3\, 82.0 & 92.2\, 92.1\, 92.3 & 79.1\, 79.8\, 79.7 & 84.6\,(+4.6) \\
\rowcolor{gray!20} ProDe-V + S\textsuperscript{2}T & \texttimes & -- & 86.1\, \textbf{86.3}\, 86.1 & 83.6\, 83.7\, \textbf{83.9} & 92.5\, 92.7\, 92.4 & 79.9\, \textbf{80.3}\, 80.0 & \textbf{85.6}\,(+0.6) \\
\midrule
\rowcolor{gray!20} ProS\textsuperscript{2}T (Ours) & \textbf{\checkmark} & -- & 86.0 & 83.6 & \textbf{92.8} & 80.0 & \textbf{85.6} \\
\bottomrule
\end{tabular}
\end{adjustbox}
\end{table*}

Tables~\ref{tab:office_home_visda} and~\ref{tab:domainnet126} compare ProS\textsuperscript{2}T with existing methods on Office-Home, VisDA, and DomainNet-126. Despite requiring no source-trained model, ProS\textsuperscript{2}T remains competitive across all three benchmarks. On Office-Home, it reaches 85.4\%, slightly exceeding VSFOT (85.3\%), and achieves the best Product accuracy of 92.8\%. On VisDA, it obtains 91.1\% mean per-class accuracy, compared with 91.4\% for DUET and 91.0\% for ProDe-V. It also achieves the best reported result on Horse. On DomainNet-126, ProS\textsuperscript{2}T reaches 85.6\%, exceeding ProDe-V by 0.6\%.
Compared with CLIP zero-shot, ProS\textsuperscript{2}T improves average accuracy by 9.1\%, 5.8\%, and 7.3\% on Office-Home, VisDA, and DomainNet-126, respectively. 

Even compared with enhanced SFDA methods with plug-and-play S\textsuperscript{2}T integration, ProS\textsuperscript{2}T outperforms DIFO-V + S\textsuperscript{2}T and delivers comparable performance to ProDe-V + S\textsuperscript{2}T, supporting the feasibility of VF-SFDA.

Tables~\ref{tab:office_home_visda} and~\ref{tab:domainnet126} also show the results of CLIP zero-shot + S\textsuperscript{2}T, DIFO-V + S\textsuperscript{2}T, and ProDe-V + S\textsuperscript{2}T. Without training, S\textsuperscript{2}T improves CLIP zero-shot accuracy by 2.7\%, 0.5\%, and 2.5\% on Office-Home, VisDA, and DomainNet-126, respectively. On Office-Home, DIFO-V + S\textsuperscript{2}T reaches 84.5\%, matching DIFO++, while ProDe-V + S\textsuperscript{2}T achieves the highest accuracy of 85.7\%. On DomainNet-126, S\textsuperscript{2}T improves DIFO-V and ProDe-V by 4.6\% and 0.6\%, respectively. On VisDA, DIFO-V improves from 90.3\% to 90.9\%. The gains vary across methods and datasets.

\subsection{Ablation Study}\label{sec:ablation}

\begin{table}[t]
\centering
\renewcommand{\arraystretch}{1.08}
\setlength{\tabcolsep}{2pt}
\fontsize{8.27}{9.95}\selectfont
\caption{Component ablations (\%). \textbf{bold}: best.}
\label{tab:ablation}
\begin{tabularx}{\columnwidth}{Xccc}
\toprule
Variant & \shortstack{Office-\\Home} & \shortstack{DomainNet-\\126} & Avg. \\
\midrule
\rowcolor{gray!20} \textbf{Full} & \textbf{85.4} & \textbf{85.6} & \textbf{85.5}\,(+0.0) \\
\midrule
\shortstack[l]{w/o S\textsuperscript{2}T} & 83.8 & 84.1 & 84.0\,(-1.5) \\
Fixed $\lambda=1$ & 84.1 & 83.9 & 84.0\,(-1.5) \\
\shortstack[l]{w/o logit fusion} & 84.2 & 83.9 & 84.1\,(-1.4) \\
w/o Prompt Tuning & 82.8 & 84.4 & 83.6\,(-1.9) \\
\midrule
w/o IIC & 84.2 & 84.6 & 84.4\,(-1.1) \\
w/o Diversity & 84.5 & 84.4 & 84.5\,(-1.0) \\
w/o Hard CE & 5.7 & 0.8 & 3.3\,(-82.2) \\
\bottomrule
\end{tabularx}
\end{table}

We conduct ablation experiments on Office-Home and DomainNet-126 to evaluate the effectiveness of each key component. Table~\ref{tab:ablation} reports the results. The full model (Full) achieves an average accuracy of 85.5\% across the two datasets, serving as the reference for comparison.

First, we remove S\textsuperscript{2}T and generate pseudo-labels using only the main view (w/o S\textsuperscript{2}T). The average accuracy decreases to 84.0\%, a drop of 1.5\%, showing the benefit of complementary views for pseudo-label correction. Fixing $\lambda=1$ from the start reduces average accuracy to $84.0\%$ (-1.5\%), suggesting that unreliable early target predictions can degrade supervision.
Removing logit fusion (w/o logit fusion) reduces average accuracy to $84.1\%$ (-1.4\%), showing the benefit of combining ViL predictions with the target model's learned information. Keeping CLIP prompts fixed (w/o Prompt Tuning) lowers average accuracy to $83.6\%$ (-1.9\%), highlighting the benefit of adapting ViL guidance to the target task.

We further remove individual loss terms to examine their contributions. Without mutual-information alignment (w/o IIC), the average accuracy falls to 84.4\%, a drop of 1.1\%. 
Removing diversity regularization (w/o Diversity) reduces the average accuracy to 84.5\%, a drop of 1.0\%, supporting the role of class balance in avoiding biased predictions.
Removing hard pseudo-label cross-entropy (w/o Hard CE) reduces average accuracy to $3.3\%$, indicating that learning from random initialization relies heavily on hard pseudo-label supervision.

\subsection{Analysis and Validation of S\textsuperscript{2}T}\label{sec:view_analysis}

\begin{figure}[h]
\centering
\begin{subfigure}[t]{0.326\columnwidth}
\centering\includegraphics[width=\linewidth]{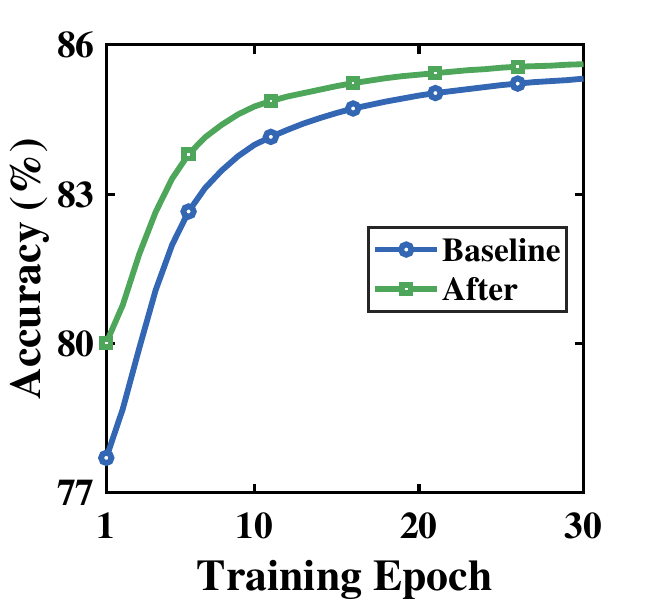}\caption{}\label{fig:correction_a}
\end{subfigure}\hfill
\begin{subfigure}[t]{0.326\columnwidth}
\centering\includegraphics[width=\linewidth]{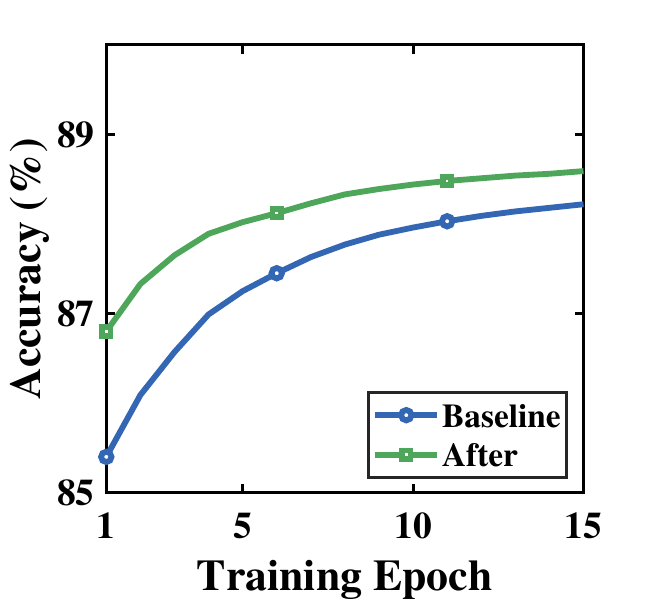}\caption{}\label{fig:correction_b}
\end{subfigure}\hfill
\begin{subfigure}[t]{0.326\columnwidth}
\centering\includegraphics[width=\linewidth]{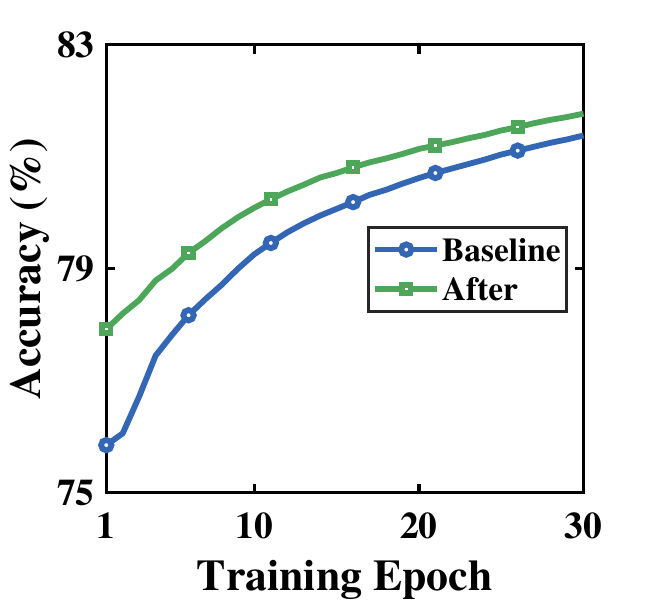}\caption{}\label{fig:correction_c}
\end{subfigure}
\caption{Hard pseudo-label accuracy before (Baseline) and after S\textsuperscript{2}T on three adaptation tasks: (a) Office-Home $\rightarrow$A. (b) VisDA-C. (c) DomainNet-126 $\rightarrow$C.}
\label{fig:correction_curves}
\end{figure}

Figure~\ref{fig:correction_curves} shows that S\textsuperscript{2}T consistently improves the pseudo-label accuracy throughout training. 
The larger early gain is consistent with stronger reliance on ViL guidance when the target model is still weak. Single-view errors therefore have a greater influence on supervision, making complementary views particularly valuable. As the target model matures, the main-view fusion becomes more reliable and the gain narrows, but remains positive.

Table~\ref{tab:complementary_view_controls} compares different spatial views and selection strategies on Office-Home and DomainNet-126. Structure Priority and Shape Priority achieve average accuracies of 84.0\% and 84.1\%, respectively. Their comparable performance suggests that both image coverage and shape preservation provide useful recognition cues.

Repeated Structure Priority uses two independently main views and reaches 84.3\%, slightly above Structure Priority alone, suggesting that repeating the same transformation type provides limited additional evidence. 
Randomly choosing between Structure Priority and Shape Priority reaches only $84.3\%$, indicating limited gains from transformation diversity alone. Two-view Averaging and ViL-only Selection achieve higher accuracies of $85.0\%$ and $85.1\%$, respectively, showing the benefit of using evidence from both views. However, the former averages the fused distributions without explicitly assessing candidate reliability, while the latter selects a fused candidate using ViL entropy before fusion, excluding target-model information from the confidence comparison. Our method selects the more confident candidate after fusion and reaches $85.5\%$, the best result among non-oracle variants.

Student-side Selection applies confidence-based selection to target model predictions while keeping the ViL view fixed. It achieves 84.2\%, 1.3\% below ours, supporting that pretrained ViL predictions can provide more reliable candidates.

Finally, the oracle reference uses target labels to select the correct candidate whenever one is available. It achieves 87.5\%, exceeding our method by 2.0\%. This gap indicates that the two views contain useful complementary evidence that the current confidence measure does not fully exploit, leaving substantial research value.

\begin{table}[t]
\centering
\renewcommand{\arraystretch}{1.08}
\setlength{\tabcolsep}{2pt}
\fontsize{8.27}{9.95}\selectfont
\caption{Spatial views and selection strategies (\%). \textbf{Bold}: best non-oracle results. Oracle uses target labels.}
\label{tab:complementary_view_controls}
\begin{tabularx}{\columnwidth}{Xccc}
\toprule
Variant & \shortstack{Office-\\Home} & \shortstack{DomainNet-\\126} & Avg. \\
\midrule
Structure Priority only & 83.8 & 84.1 & 84.0\,(-1.5) \\
Shape Priority only & 84.1 & 84.0 & 84.1\,(-1.4) \\\hline
\shortstack[l]{Repeated Structure Priority}
& 84.2 & 84.4 & 84.3\,(-1.2) \\
Random Structure/Shape & 84.1 & 84.4 & 84.3\,(-1.2) \\
Two-view Averaging & 84.9 & 85.0 & 85.0\,(-0.5) \\
ViL-only Selection & 85.0 & 85.1 & 85.1\,(-0.4) \\\hline
Student-side Selection & 84.0 & 84.3 & 84.2\,(-1.3) \\
\midrule
\rowcolor{gray!20}
\textbf{Ours} & \textbf{85.4} & \textbf{85.6}
& \textbf{85.5}\,(+0.0) \\
\midrule
Oracle & 87.1 & 87.8 & 87.5\,(+2.0) \\
\bottomrule
\end{tabularx}
\end{table}

\subsection{Validation of the Adaptation Dynamics}\label{sec:dynamic_validation}

We examine the adaptation dynamics by tracking randomly initialized (Random) and source-initialized (Source-C) models on Office-Home Cl\(\to\)Ar. We measure the mean normalized Jensen--Shannon divergence (JSD)~\cite{lin1991divergence}  on the same whole target samples and examine three aspects of the adaptation process.

\begin{figure}[t]
\centering
\begin{subfigure}[t]{0.326\columnwidth}
\centering\includegraphics[width=\linewidth]{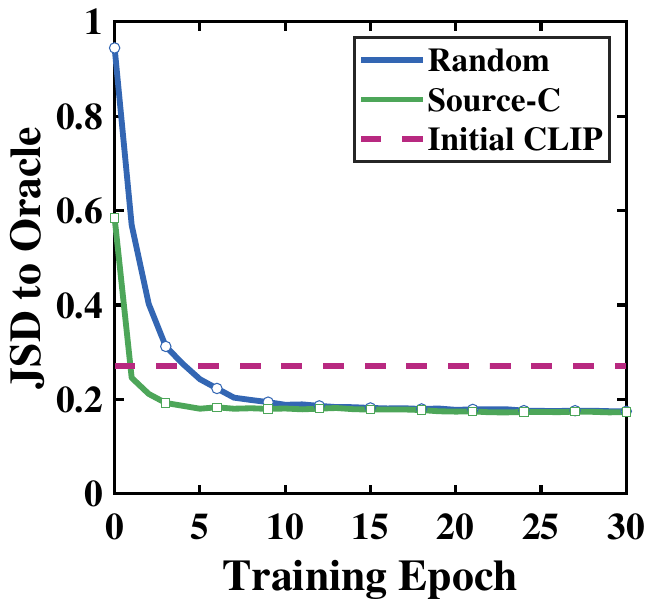}\caption{}\label{fig:jsd_oracle}
\end{subfigure}\hfill
\begin{subfigure}[t]{0.326\columnwidth}
\centering\includegraphics[width=\linewidth]{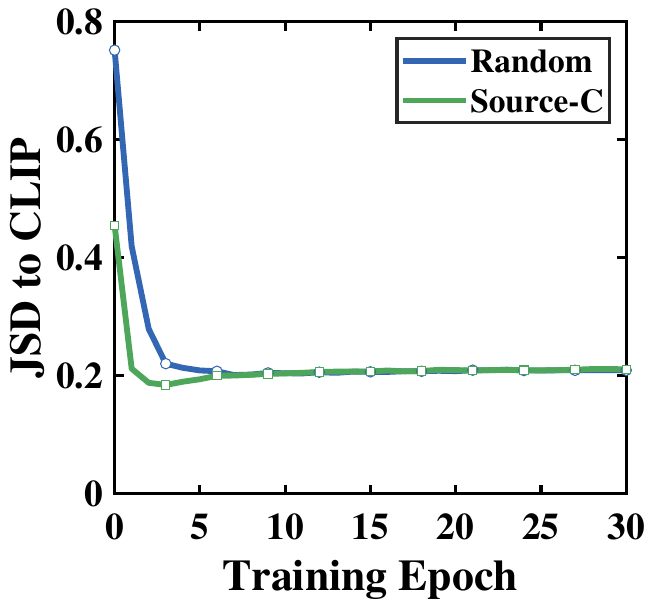}\caption{}\label{fig:jsd_clip}
\end{subfigure}\hfill
\begin{subfigure}[t]{0.326\columnwidth}
\centering\includegraphics[width=\linewidth]{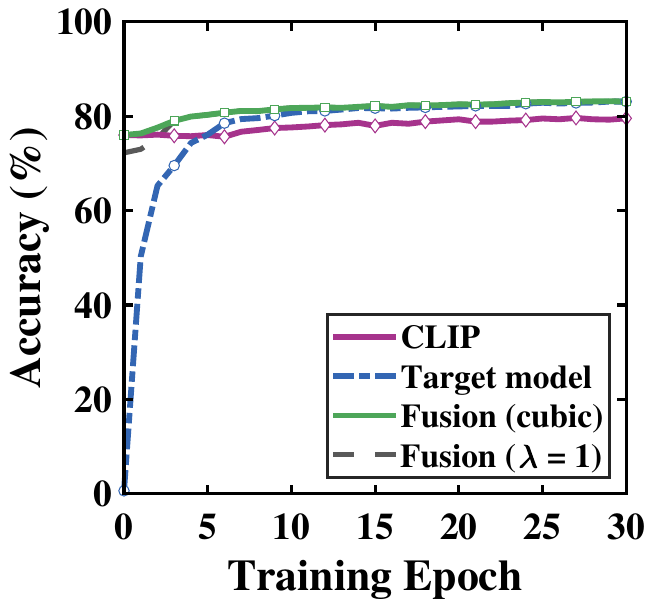}\caption{}\label{fig:fusion_accuracy}
\end{subfigure}
\caption{Adaptation dynamics on Office-Home. (a) JSD to the oracle, with initial CLIP as a fixed baseline. (b) JSD to the initial CLIP reference. (c) Accuracy of CLIP, the target model, and fused predictions. Fixed-$\lambda$ fusion is evaluated at the same progressively trained checkpoints.}
\label{fig:dynamic_validation}
\end{figure}

Fig.~\ref{fig:dynamic_validation}\subref{fig:jsd_oracle} compares both models with one-hot oracle distributions constructed from target labels, representing \(D_t\) in Fig.~\ref{fig:dynamic_process}. The initial CLIP-to-oracle JSD serves as a fixed baseline. Initially, CLIP is closer to the oracle than either model. During adaptation, both models substantially reduce their JSD and reach similar final levels below this baseline. The rapid initial decrease shows that both models approach their late-stage JSD levels early in training. These trends provide empirical support for the adaptation process illustrated in Fig.~\ref{fig:dynamic_process}.

Fig.~\ref{fig:dynamic_validation}\subref{fig:jsd_clip} uses the pre-adaptation CLIP predictions as a fixed reference. Random initially has a much larger JSD than Source-C, but both curves decrease rapidly and settle at similar levels relative to this shared reference.

Fig.~\ref{fig:dynamic_validation}\subref{fig:fusion_accuracy} compares the accuracies of CLIP, the target model, and their fused predictions. Early in training, fixed-weight fusion (\(\lambda=1\)) performs worse than CLIP, whereas progressive fusion limits the influence of unreliable target predictions. The target model improves rapidly under ViL guidance and eventually surpasses CLIP. At later stages, fusion achieves slightly higher accuracy than either model alone. These results are consistent with Fig.~\ref{fig:dynamic_process}: the target model first learns under ViL guidance, and its late-stage predictions \(D_i^{\mathrm{late}}\) then complement the ViL predictions \(D_v\), yielding more accurate fused supervision \(D_f\).

\section{Conclusion}
Most ViL-guided SFDA methods depend on source-trained models. This paper revisits this dependence through VF-SFDA, where a target model learns from random initialization using only unlabeled target data and a pretrained ViL model. Our analysis of adaptation dynamics highlights the importance of reliable ViL supervision. However, spatial preprocessing can weaken this supervision by distorting object proportions or removing important structural cues. To address this issue, we propose ProS\textsuperscript{2}T. Its S\textsuperscript{2}T module pairs Structure Priority and Shape Priority views to provide complementary class evidence. Progressive fusion gradually incorporates learned target-domain information while limiting unreliable early target predictions, and confidence-based selection refines hard pseudo-labels. Experiments on Office-Home, VisDA-C, and DomainNet-126 demonstrate competitive adaptation performance without source models. S\textsuperscript{2}T also serves as a plug-and-play module that can be used in ViL-guided SFDA methods.

\Needspace{6\baselineskip}
\section*{LLM Usage Statement}
ChatGPT was used to assist with English drafting and language editing, including the abstract and conclusion, literature lookup, plotting and analysis code support, and LaTeX preparation. The authors remain responsible for all scientific claims, experimental results, and references.

{\small
\balance
\bibliographystyle{ieeenat_fullname}
\bibliography{references}
}

\end{document}